# GENETIC ALGORITHMS FOR THE RESOURCE-CONSTRAINED PROJECT SCHEDULING PROBLEM IN AIRCRAFT HEAVY MAINTENANCE

**Kusol Pimapunsri[1,*], Darawan Weeranant[2] and Andreas Riel[3]**

## Abstract

**Due to complex sets of interrelated activities in aircraft heavy maintenance (AHM), many airlines have to deal with substantial aircraft maintenance downtime. The scheduling problem in AHM is regarded as an NP-hard problem. Using exact algorithms can be time-consuming or even infeasible. This article proposes genetic algorithms for solving the resource-constrained project scheduling problem (RCPSP) in AHM. The objective of the study was to minimise the makespan of the maintenance plan. The proposed algorithms applied five heuristic dispatching rules to generate an initial population based on activity list formation. Resource allocation methods for RCPSP – earliest start time (EST) and workgroup and earliest start time (WEST) – were used to evaluate the fitness value. The elitist and roulette wheel methods were applied in the selection process. The sequences of the activity lists were then iteratively improved by crossover and mutation operations. The results show that the proposed algorithms perform efficiently compared to the existing solutions in terms of computational time and resource allocation.**

## Introduction

The demand for commercial air transport has increased over the last few decades due to short travel times and affordable prices. The annual growth in passenger traffic from 2015 to 2034 is estimated to be 4.6-4.9% (Schmidt, 2017). This rapid growth has imposed many challenges for planning, scheduling, and operations in the aviation industry.

Aircraft maintenance, repair, and overhaul (MRO) activities are critical for aircraft safety, and periodic maintenance checks need to be carried out on each aircraft after a specified number of flying hours. The cost of MRO is about 9% of the annual operating costs of airlines, and it is the third highest after fuel and labour costs (Qin *et al.*, 2019). Furthermore, airlines cannot generate income during maintenance periods. Therefore, they wish to minimise maintenance lead-time while assuring that all operations are in compliance with the quality standards and regulatory requirements of Airworthiness Authorities, such as the Federal Aviation Administration (FAA), European Aviation

[1]  Department of Industrial Engineering, Faculty of Engineering, King Mongkut's University of Technology North Bangkok, Bangkok, Thailand. E-mail: kusol.p@eng.kmutnb.ac.th
[2]  Department of Industrial Engineering, Faculty of Engineering, Rajamangala University of Technology Phra Nakhon, Bangkok, Thailand.
[3]  G-SCOP Laboratory, Grenoble Institute of Technology, Grenoble, France.
*   Corresponding author

Safety Agency (EASA), and International Civil Aviation Organization (ICAO).

Aircraft maintenance scheduling is a complicated task involving the synthesis of a range of economic, political, legal and technical factors. The demands on service, aircraft use and operational costs of the aircraft are the principal drivers of this problem (Sriram and Haghani, 2003). In particular, aircraft heavy maintenance (AHM), also called a heavy maintenance visit, makes the maintenance scheduling problem even more difficult to solve. AHM consists of complex sets of interrelated activities that must be performed within a given period of time. They usually require keeping the aircraft in a hangar for several weeks to carry out inspection and maintenance activities that are compliant with specific maintenance programmes. Due to the enormous number of interrelated activities in AHM, it is challenging for the planner to minimise the makespan of the maintenance plan with scarce resources, while ensuring that the aircraft is delivered on time or with the shortest possible delay. This problem, the so-called resource-constrained project scheduling problem (RCPSP), is regarded as a nondeterministic polynomial (NP)-hard problem in AHM. Using exact algorithms for solving it leads to extremely long computation times for obtaining the optimal solution, and may even be intractable.

In this paper, we present genetic algorithms for solving the RCPSP in AHM. The scope of the problem we dealt with in this study was a heavy maintenance visit (D check) of an aircraft cabin, which is composed of five main workgroups, i.e. cockpit, door, galley, interior, and lavatory. Only scheduled maintenance tasks were considered, while unexpected or unscheduled maintenance requirements were not taken into account. The objective of this study was to develop genetic algorithms to evaluate solutions with a minimum makespan of the maintenance plan in reasonable computational time.

## Aircraft Heavy Maintenance (AHM)

Civil aircraft maintenance can be subdivided into four major types of investigations, i.e. from A check through to D check. The type of check depends on the number of hours the aircraft has flown since its last check, the age of the aircraft, and the number of take-off and landing cycles it carried out. A and B checks are lighter checks, while C and D checks are considered heavier checks which must be performed inside a hangar using specialized equipment and highly trained personnel (Friend, 1992). The C and D checks are considered heavy maintenance. The C check typically occurs every two years and requires 10,000-30,000 man-hours (2 to 4 weeks) to inspect and exhaustively overhaul the entire aircraft. The D check, the most comprehensive and occurring approximately every six years, involves taking the entire airplane apart for inspection and overhaul. It needs up to 50,000 man-hours to complete (UK Department for Business, Innovation and Skills, 2016). However, it may require a different number of resources depending on the size and complexity of the aircraft, the guidelines of aircraft manufacturers, and the requirements of the airline. Our previous study (Weeranant and Pimapunsri, 2017) examined the complexity of AHM by applying a design structure matrix (DSM; Yassine, 2004; Eppinger and Browning, 2012) and simulation technique to ascertain the proper sequence for the aircraft maintenance plan.

## The Resource-Constrained Project Scheduling Problem

The RCPSP is a classical problem that has received the attention of many researchers for several decades. The objective of RCPSP is to properly schedule dependent activities over time such that the makespan of the project is minimised while precedence and resource constraints are met (Wang, 2016; Kolisch and Hartmann, 2017). The project scheduling problem is identified as determining the time required to implement the activities of a project to achieve a certain objective (Habibi *et al*., 2018).

Heuristic scheduling is one of the traditional methods for seeking solutions close to the optimal with acceptable computational cost and usually requiring less time. The heuristics are often defined as scheduling rules with dispatch rules. Some heuristic algorithms (Tormos and Lova, 2001; Sriram and Haghani, 2003; Rosales, 2015; Kolisch and Hartmann, 2017) have been developed to solve the RCPSP, as well as a simulation approach (Pimapunsri and Weeranant, 2018) for examining uncertainty in activities and delays in AHM. However, there are still no promising methods that guarantee optimal solutions and computational feasibility. Although the heuristic methods provide good solutions in reasonable computation time, they are typically limited to a specific set of constraints or problem formulation. In addition, development of new heuristics is difficult.

In general, a large-scale scheduling problem is NP-hard. Algorithms exist for exactly solving some forms of the problem, but they typically take too long when the size of the problem grows or additional constraints are added. As a result, most research has been devoted to either simplifying the scheduling problem to the point where some algorithms can find solutions, or to devising efficient heuristics for finding good solutions.

## Genetic Algorithms

Genetic algorithms (GAs) form a subset of stochastic search optimisation methods introduced in the mid-1970s by Holland (Holland, 1975). Based on the principles of natural evolutionary processes, GAs create new solutions by employing heuristic search methods, such as selection, crossover, and mutation, to find better solutions (Goldberg, 1989; Michalewicz, 1994). GAs are widely used in various engineering applications, e.g. production scheduling (Goncalves *et al.*, 2005, 2008) and design optimisation (Gokdag and Yildiz, 2012; Kiani and Yildiz, 2016), and in solving RCPSP (Wall, 1996; Franco *et al.*, 2007; Kadam and Kadam, 2014). RCPSP consists of several candidate solutions. In GAs, each solution is called an individual (chromosome) which has a set of properties (its genes) that can be mutated to generate the next-generation population of solutions. A fitness function is used in the evaluation to select the best individuals. As the GA is a heuristic search method based on the survival of the fittest on optimisation, it is an effective method for solving RCPSP. The procedures and the pseudo-code of the proposed algorithm are shown in Figures 1 and 2.

## Research Methodology

### Chromosome Structure

Usually, several hundred activities have to be completed to accomplish the heavy maintenance of an aircraft cabin. In this study, we defined an activity list as a chromosome. A chromosome is composed of genes; the number given to each gene corresponds to the ID number and its information (the duration and resources required). Figure 3 provides an example of a network of ten activities. This network can create 10! chromosomes, without taking into account precedence constraints. For example, a chromosome of this network could be [5-6-4-10-9-1-3-7-2-8].

### Initial Population

The population of solutions is based on activity list formation. An activity list is composed of activities according to their precedence in the relationship. A highly constrained scheduling problem regularly has a small feasible search space. Purely random generation of activity sequences results in a large number of infeasible solutions. In this study, we used a permutation-based simulation procedure to produce an initial population of precedence-feasible chromosomes (Hartmann and Drexl, 1998). The permutation encoding method creates an initial population based on its size. The objective of this step is to determine a set of activity lists (initial population) by using five dispatching rules: SPT (shortest processing time); LPT (longest processing time); critical path and SPT; critical path

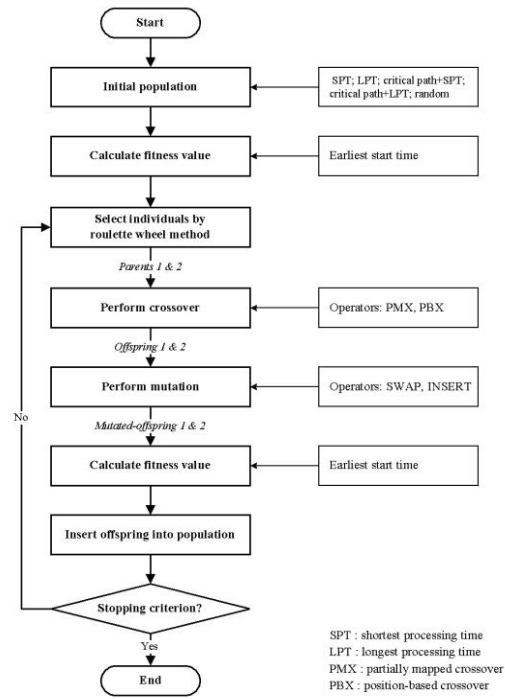

Figure 1. Flowchart of the proposed genetic algorithm

```
overall procedure: GA algorithm for task scheduling
input: job constraints, resource information, GA parameters
output: solution from the GA algorithm
begin
        n←0; //n: generation number
//Step 1:  Construct initial population
        create Population(n) consists of Y chromosomes; //Y: population size
        repair Population(n); repair Y chromosomes based on task constraints
        while (not terminating condition of GA)
        do
//Step 2:  Evaluate performance of each chromosome
        Each chromosome is evaluated by the makespan
//Step 3:  Selection
        apply elitist method to Population(n);
        select Parent(n) from Population(n) by roulette wheel method;
//Step 4:  Apply genetic operators
        crossover Parent(n) to get Offspring(n);
        mutate Offspring(n) to get MOffspring(n); //MOffspring: mutated offspring
        repair MOffspring(n);
        Population(n+1)←a sequence from elitist method and MOffspring(n);
        calculate makespan of Population(n+1);
        n←n+1;
    end;
    select the sequence with minimum makespan from Population(n);
output: solution from the GA algorithm
end;
```

Figure 2. The pseudo-code of the proposed genetic algorithm

and LPT; and random rules. The number of activity lists generated depends on the required population size. For example, if the population size is set to 10, this method creates two equal chromosomes (activity lists) for each dispatching rule. However, only chromosomes that meet precedence constraints are considered. Otherwise, they need to be repaired before being evaluated. According to the chromosome, [5-6-4-10-9-1-3-7-2-8] in the previous step is not a proper sequence for the precedence constraints in Figure 3. Then, a repair process is carried out by exchanging the position of activities (genes) of the same priority and a new chromosome is formed in the population, for example [5-6-4-2-8-1-7-9-3-10].

**Evaluation of Fitness Value**

In each generation, all chromosomes in the population are evaluated by the fitness function. The chromosomes with better fitness values are included in the mating pool to form new offspring (child chromosomes). The fitness function ($F_i$) of the proposed GA is defined in Equation (1). The fitness value represents the makespan ($C_{max}$). The objective function is to minimise the makespan. Then, the chromosome that has the lowest value is considered the best current solution. Selection of chromosomes with better fitness value enables the GA to reach solutions in a shorter period of time.

$$F_i = Minimize\ C_{max} \qquad (1)$$

A chromosome is an activity list formation which is a representation of the schedule. In this step, RCPSP is used to schedule the resource demand of each activity for the sequences and resource allocation methods. To evaluate the fitness value of each chromosome, we applied the earliest start time (EST) algorithm since it is simple and efficient and is widely used to minimise the makespan. Furthermore, the workgroup of the activity was considered for scheduling resources. Therefore, the workgroup and earliest start time (WEST) algorithm was also applied to evaluate the fitness value in this study.

EST is one of the greedy algorithms that are used to solve scheduling problems. According to the sequences of each activity list (chromosome) generated from the previous step, EST first considers the resource which prompts the activity. In case the number of available resources is higher than demand, EST takes the resource prior to the lowest identification number. This process is repeated until all the activities are accomplished, then the duration (makespan) of each chromosome is calculated. The schedule of the chromosome [5-6-4-2-8-1-7-9-3-10] can be represented as in Figure 4.

The WEST algorithm, according to EST, takes into account the workgroup to which the activity under consideration belongs. Aircraft maintenance (assembly and disassembly) activities are mostly operated in the hangar. Therefore, the resources required are moved to the hangar. When the activity considered is in the same workgroup and requires the same resource as the previous activity, the resource that operates the previous activity is assigned first. The reason is to reduce the moving time of resources and equipment between workgroups that are usually in different locations. For example, the schedule of the same chromosome [5-6-4-2-8-1-7-9-3-10] would be represented in a different way as shown in Figure 5.

**Selection Operation**

The objective of this step is to select parent chromosomes (sequences of activity lists) obtained from the previous step. This study applied two selection methods: called elitist and roulette wheel (Pasala *et al.*, 2013). The elitist method selects the sequence of activity lists that offers the minimum fitness value, while the roulette wheel method is a probabilistic selection. The reciprocal of the fitness value and the probability of each sequence of activity lists are calculated by Equations (2) and (3), respectively. Then, the parent sequences of activity lists are selected based on the probability interval of each sequence of activity lists by using random numbers. This method uses the strength of each chromosome to create area proportions (probability) on the wheel. Let $h$ be the sequence index of the activity list starting from 1 to $Ps$ (population size) and $R_h$ and $P_h$ be the reciprocal of the fitness value and the probability of sequence $h$:

$$R_h = \frac{1}{F_h} \quad (2)$$

$$P_h = \frac{R_h}{\sum_{h=1}^{Ps} R_h} \quad (3)$$

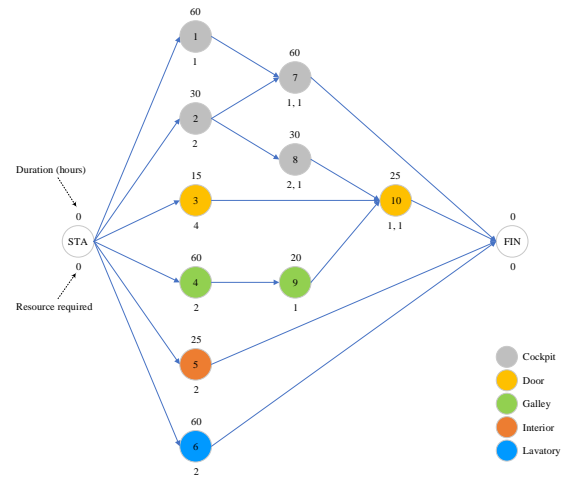

**Figure 3. An example of a network information**

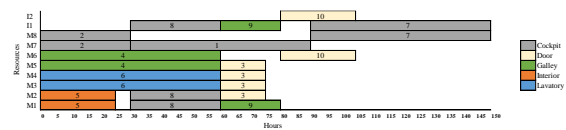

**Figure 4. Schedule of operations of the EST algorithm**

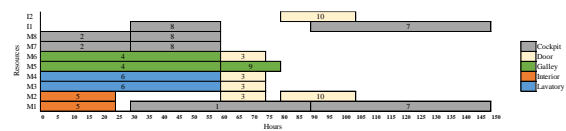

**Figure 5. Schedule of operations of the WEST algorithm**

**Crossover Operation**

Two individuals, denoted as parent chromosomes, are selected from the previous step. To produce a new generation, their genes are exchanged in a certain order to obtain offspring (new child) chromosomes. There are several crossover operators in the literature for creating new child chromosomes. In this study, we applied two well-known crossover operators, i.e. partially mapped crossover (PMX) and position-based crossover (PBX), since they always offer good offspring for permutation encoding (Sukkerd and Wuttipornpun, 2016). Note that the crossover process is allowed when a random number is less than the crossover probability ($Pc$).

PMX is one of the most popular and effective crossovers for order-based GA to deal with combinatorial optimisation problems (Ting *et al.*, 2010). Duplicate numbers in the offspring chromosomes are not allowed. To address the issue of the legality of an order, PMX uses a mapping relationship to legalise the offspring that have duplicate numbers. Figure 6 illustrates how PMX legalises offspring. First, two points on Parent 1 and Parent 2 that will create three substrings are marked as shown in Step-1. Substring 2 (4-2-8-1-7) of Parent 1 and (6-2-5-3-7) of Parent 2 are then exchanged to produce proto-offspring in Step-2. However, this results in duplication of the genes (5, 6, 3) that appear twice in proto-offspring 1 as well as the genes (1, 4, 8) in proto-offspring 2. To fix the illegal offspring, Step-3 partially maps the genes in substring 2, e.g. '6' to '4', '5' to '8', and '3' to '1'. Then, the duplicated genes in substring 1 and substring 3 are replaced with the corresponding genes in the mapping relationship as shown in Step-4.

PBX was proposed by Syswerda (1991). First, a random set of positions is selected in the parent strings, for example, a pair of parents as shown in Figure 7. Suppose that the second, fifth and eighth positions are selected as shown in Step-1. The first offspring's positions are filled by selecting the same positions of Parent 1. Then deselected positions of Parent 2 are taken in order from left to right without duplicate number. The same procedure is applied to the second offspring. The results are shown in Step-2.

**Mutation Operation**

The mutation operator is applied to prevent the offspring sequences from falling into a local optimum area. Two mutation methods, SWAP and INSERT, were applied in this study since they are very efficient for permutation encoding. These random operators enable individuals to make a new combination. Similar to the crossover process, the mutation process is allowed when a random number is less than the mutation probability ($Pm$; Chen *et al.*, 2013). However, similar to the initial step, the sequences of the new combination must respect the precedence constraints. Otherwise, they need to be repaired before the fitness value is evaluated. Then, the existing population is replaced by a new generation obtained by this step. This is followed by checking the stopping criterion. If it is not satisfied, a new loop from the selection operation process is started to find a new generation of the population. For example, suppose that mutation probability ($Pm$) is 0.01 and the random numbers of the two offspring in Figure 6, [8-4-6-2-5-3-7-9-1-10] and [3-6-4-2-8-1-7-5-9-10], are 0.5 and 0.004. Since the random number of the second offspring is less than $Pm$, the sequence of this offspring is mutated. In the SWAP mutation method, two genes on the chromosome are selected at random and exchange the values, while the INSERT mutation method moves a randomly chosen gene to another randomly chosen position, as shown in the examples in Figures 8 and 9, respectively.

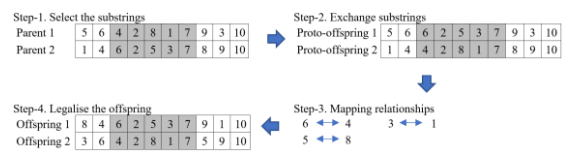

**Figure 6.** Example of partially mapped crossover (PMX)

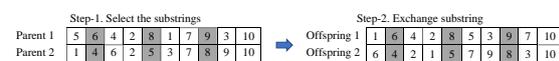

**Figure 7.** Example of position-based crossover (PBX)

## Experiment and Analysis

The experiments described in this section used the MATLAB programme on a desktop computer with an Intel® Core$^{TM}$ i5 Duo Processor 3 GHz and 4 GB RAM. The AHM case study was the same as in our previous work (Weeranant and Pimapunsri, 2017). The network information was rearranged following the precedence constraints using the design structure matrix (DSM) method (Yassine and Braha, 2003; Yassine, 2004). The characteristics of the case study were as follows: the number of activities was 317; the number of workgroups was 5; and the number of resource groups was 12. This experiment was conducted based on a factorial design. The stopping criterion was computational time which was 120 minutes maximum. There were five parameters with two algorithms, in which the total runs were $(5 \times 2 \times 3 \times 2 \times 3) \times 2$ algorithms = 360 runs which required 720 h. The parameters used in this experiment are shown in Table 1.

The response variable of the experiment was the makespan obtained from each run. Table 2 illustrates the results of the minimum makespan obtained from the EST and WEST algorithms. Although both algorithms offered the same minimum makespan of 32.995 days, the EST algorithm consumed less time than the WEST algorithm in the overall results. To evaluate the performance of both algorithms in terms of computational time, the convergence curves of the EST and WEST algorithms are shown in Figures 10 and 11, respectively. The best parameter setting of the algorithms that found the minimum makespan with the least time, 4.90 min for EST and 34.31 minutes for WEST, are shaded in Table 2.

Table 3 shows the results for minimum makespan and resource demand. In the case of resource allocation, the WEST algorithm required 115 units, while the EST algorithm required 130 units. Compared to the plan estimated by the priority rule-based heuristic in Weeranant and Pimapunsri (2017) and an experienced planner, the makespan of the proposed algorithms was significantly improved by 25.7-29.8%, while the peak resource demand was improved by 4.4-15.4%.

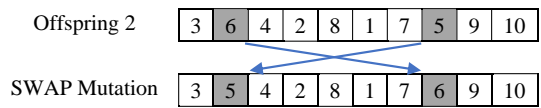

**Figure 8. SWAP mutation operator**

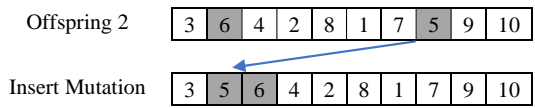

**Figure 9. INSERT mutation operator**

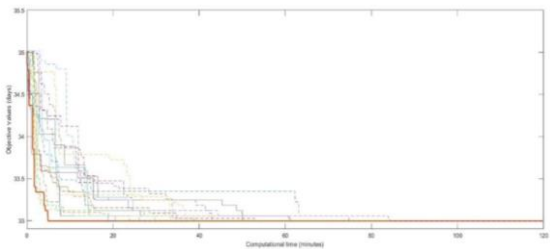

**Figure 10. Convergence curves of the EST algorithm**

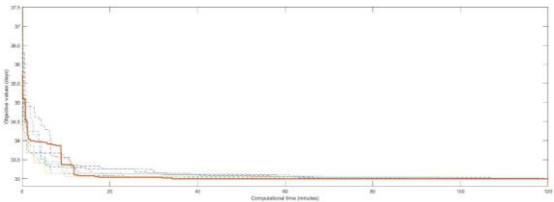

**Figure 11. Convergence curves of the WEST algorithm**

## Conclusions

This study proposed genetic algorithms to solve RCPSP in AHM. Heuristic dispatching rules were applied to determine a set of initial sequences of activity lists as an initial population. The fitness value was calculated by two algorithms of resource scheduling, EST and WEST, to obtain the makespan of each chromosome. The roulette wheel method was applied to select the parent chromosomes which subsequently evolved by crossover and mutation

**Table 1. Parameters of the experiment**

| Algorithm | Ps | Cross | Pc | Mutate | Pm |
|---|---|---|---|---|---|
| EST WEST | 5, 10, 30, 60 100 | PMX PBX | 0.7, 0.8 0.9 | SWAP INSERT | 0.05, 0.01 0.1 |

*Ps*: Population size; Cross: Crossover operator; *Pc*: Crossover probability; Mutate: Mutation operator; *Pm*: Mutation probability.

**Table 2. Best parameter settings for each algorithm**

| EST Algorithm | | | | | | WEST Algorithm | | | | | |
|---|---|---|---|---|---|---|---|---|---|---|---|
| *Ps* | *Pc* | Cross | *Pm* | Mutate | Time | *Ps* | *Pc* | Cross | *Pm* | Mutate | Time |
| 5 | 0.7 | PBX | 0.05 | SWAP | 50.02 | 5 | 0.7 | PMX | 0.1 | SWAP | 88.10 |
| 5 | 0.8 | PBX | 0.05 | SWAP | 33.15 | 5 | 0.9 | PBX | 0.1 | SWAP | 113.66 |
| 5 | 0.9 | PMX | 0.05 | SWAP | 61.21 | 10 | 0.7 | PMX | 0.01 | SWAP | 117.68 |
| **10** | **0.7** | **PMX** | **0.1** | **SWAP** | **4.90** | 10 | 0.8 | PMX | 0.1 | SWAP | 60.04 |
| 10 | 0.7 | PBX | 0.05 | SWAP | 33.39 | 30 | 0.7 | PMX | 0.05 | SWAP | 78.55 |
| 10 | 0.7 | PBX | 0.1 | SWAP | 43.02 | 30 | 0.7 | PBX | 0.1 | SWAP | 107.04 |
| 10 | 0.9 | PMX | 0.05 | SWAP | 18.82 | 30 | 0.8 | PMX | 0.01 | SWAP | 119.74 |
| 30 | 0.7 | PMX | 0.05 | SWAP | 38.90 | **30** | **0.8** | **PMX** | **0.1** | **SWAP** | **34.31** |
| 30 | 0.7 | PBX | 0.1 | SWAP | 50.46 | | | | | | |
| 30 | 0.8 | PBX | 0.05 | SWAP | 53.00 | | | | | | |
| 30 | 0.9 | PMX | 0.05 | SWAP | 20.55 | | | | | | |
| 60 | 0.7 | PMX | 0.1 | SWAP | 26.16 | | | | | | |
| 60 | 0.8 | PMX | 0.01 | SWAP | 29.75 | | | | | | |
| 60 | 0.8 | PMX | 0.1 | INSERT | 74.81 | | | | | | |
| 60 | 0.9 | PMX | 0.05 | INSERT | 23.87 | | | | | | |
| 100 | 0.8 | PMX | 0.1 | INSERT | 38.11 | | | | | | |
| 100 | 0.9 | PMX | 0.1 | INSERT | 84.37 | | | | | | |

*Ps*: Population size; *Pc*: Crossover probability; Cross: Crossover operator; *Pm*: Mutation probability; Mutate: Mutation operator.

**Table 3. Results of makespan and resource demand analysis**

| Algorithm | Makespan (days) | Max. of Resource Demand (units) |
|---|---|---|
| Master plan | 47.000 | 136 |
| Priority rule-based heuristic | 44.420 | 136 |
| Genetic algorithm: EST | 32.995 | 130 |
| Genetic algorithm: WEST | 32.995 | 115 |

processes to produce a new generation of the population. The performance of the proposed algorithms was controlled by the parameters of GA: population size, operators and their probability of crossover, and mutation processes. The results of the experiment indicate that the proposed algorithms are more efficient than the existing plans for determining project makespan and resource allocation. Due to the probabilistic nature of GA, the proposed algorithms do not guarantee optimality. However, the results obtained from the proposed algorithms are clearly shown to be a significant improvement over master plan and priority rule-based heuristic methods. Future research will focus on other types of checks and heavy maintenance visits of other aircraft which usually have more or less distinct characteristics, e.g. number of activities, number of resources and resource groups, number of workgroups, etc. This future analysis is necessary to make the algorithm applicable to a wider range of case studies.

## Acknowledgments

This research was funded by King Mongkut's University of Technology North Bangkok (KMUTNB), contract no. KMUTNB-62-DRIVE-25. The authors are grateful to KMUTNB for the financial support received to execute this work.